\documentclass{article} % For LaTeX2e
\usepackage{pap,times}

\usepackage{epsfig}
\usepackage{graphicx}
\usepackage{amsmath,amsthm,amssymb}
\usepackage{algorithm}
\usepackage{algorithmic}
\usepackage{times}

\def\beq#1\eeq{\begin{equation}#1\end{equation}}
\def\beqn#1\eeqn{\begin{displaymath}#1\end{displaymath}}
\def\beqa#1\eeqa{\begin{eqnarray}#1\end{eqnarray}}
\newcommand{\Figref}[1]{Fig.~\ref{#1}}

\newtheorem{theorem}{Theorem}

\newtheorem{corollary}[theorem]{Corollary}

\newcommand{\argmax}{{\rm arg}\max}

\newcommand{\R}{{\mathbb{R}}}
\newcommand{\Expct}{{\mathbb E}}
\newcommand{\E}{{\mathbb E}}

\newcommand{\calF}{{\cal F}}

\newcommand{\calL}{{\cal L}}

\newcommand{\calS}{{\cal S}}

\newcommand{\eps}{{\epsilon}}

\newcommand{\sumin}{{\sum_{i=1}^n}}

\newcommand{\hmu}{{\widehat\mu}}
\newcommand{\hsigma}{{\widehat\sigma}}

\newcommand{\basisX}{{\widetilde X}}

\title{Domain Adaptation: \\ Overfitting and Small Sample Statistics}

\author{
Dean Foster \\
 Department of Statistics \\
University of Pennsylvania \\
\texttt{foster@wharton.upenn.edu}
\And
Sham Kakade  \\
 Department of Statistics \\
University of Pennsylvania \\
\texttt{skakade@wharton.upenn.edu}
\And 
Ruslan Salakhutdinov \\
Brain and Cognitive Sciences and CSAIL \\ 
Massachusetts Institute of Technology \\
\texttt{rsalakhu@mit.edu}
}

\nipsfinalcopy

\begin{document}

\maketitle

\vspace{-0.15in}
\begin{abstract}
\vspace{-0.1in}
  We study the prevalent problem when a test
  distribution differs from the training distribution. 
  We consider a setting where our training set consists of a small
  number of sample domains, but where we have many samples in each
  domain. Our goal is to \emph{generalize} to a new domain. For
  example, 
  we may want to learn a similarity function
  using only certain classes of objects, but we desire that this
  similarity function be applicable to object classes not present in
  our training sample (e.g. we might seek to learn that ``dogs are
  similar to dogs'' even though images of dogs were absent from our
  training set). Our theoretical analysis shows that we can select many more features
  than domains while avoiding overfitting by utilizing
  data-dependent variance properties. We present a
  greedy feature selection algorithm based on using
  $T$-statistics. Our experiments validate this theory showing that
  our $T$-statistic based greedy feature selection 
  is more robust at avoiding
  overfitting than the classical greedy procedure.
\end{abstract}

\vspace{-0.1in}
\section{Introduction}
\vspace{-0.1in}
The generalization ability of most modern machine learning algorithms
are predicated on the assumption that the distribution over training examples
(roughly) matches the distribution over the test data. There is growing
literature studying settings where this implicit assumption fails to
hold --- often referred to as domain adaptation or transfer
learning.  This problem is central in
fields such as speech recognition~\cite{legetter95}, computational
biology~\cite{liu08}, natural language
processing~\cite{blitzer06,daume07,guo09}, and web
search~\cite{chen08,gao09}.

We examine how severe this problem can be, even on one of the most
conventional benchmark datasets, the MNIST digits dataset. Here, state-of-the-art
algorithms reliably obtain classification error rates below
$1\%$, when recognizing one digit vs. the other digits.  Consider a
natural modification of this setting where we train a model to recognize
the digit ``$2$'' vs. the other \emph{even} digits.
%Here, we
%might hope that,
If we learn to recognize a ``$2$'' accurately
(vs. only even digits), then we may hope that our classifier will robustly recognize a
``$2$'' against new odd digits. Unfortunately, this is far from being true:
a logistic regression algorithm, trained on this dataset and
achieving a (true) test error rate of about $0.5\%$ (against even digits),
jumps to $35\%$ error rate when tested vs. odd digits,
a startling $7000\%$ increase
in error. While the present work uses deep belief network features \cite{HOT},
trained on unlabeled data, this situation is generic across many other common
training methods we have tried: SVMs with various kernels and
logistic/linear regression with various feature choices (where error
rates increase from hundreds to thousands of percent depending on the
details of the experiment).
%The striking issue is that the true
%performance on the training (source) distribution is not at all reflective of
%the performance on the test (target) distribution --- raising the question of how
%to control for overfitting.

\begin{figure*}[t!]
\vspace{-0.35in}
\hbox{ \centering \hspace{-0.05in}
\centerline{\includegraphics[width=5.90in,height=1.75in]{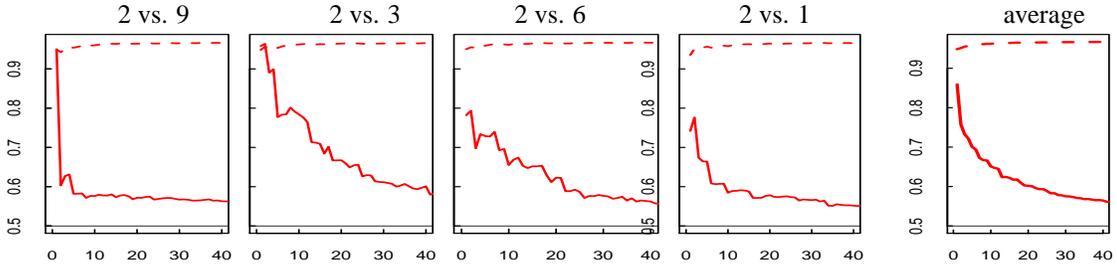}}
\begin{picture}(0,0)(0,0)
\setlength{\epsfxsize}{1.9in}
\put(-380, +108){ 2 vs. 9}
\put(-300, +108){ 2 vs. 3}
\put(-230, +108){ 2 vs. 6}
\put(-145, +108){ 2 vs. 1}
\put(-45, +108){ average }
\end{picture}
}
\vspace*{-.35in}
\caption{
\label{fig:example}
\small Area under ROC (AUROC) (y-axis) for predicting digit '2'
using the greedy
 algorithm.
The x-axis shows the order of variables
 were picked out of 2000 total variables. The top dashed
line shows generalization or 'test' performance on the source domain.  
The transfer to a new domain is
 shown below it as the solid line.  The last figure on the right shows
the average of all the ROC curves. The horizontal line is at
``chance.''}
\vspace{-0.10in}
\end{figure*}

We elucidate this overfitting issue by examining how various
``area under the ROC curves'' change as we greedily add more
features. Here, we train our model to recognize the digit ``$2$'' vs. eight other
digits, and test recognition of a ``$2$" vs. the remaining digit (with
balanced distributions where a ``$2$" appears half the time in both
the training and test distributions).
%The leftmost plot in Figure~\ref{fig:overfitting}
%shows the ROC performance on our training data as we greedily add more
%features. 
%Here, the BLAH line shows the training error and the BLAH
%line shows the hold our error (for the \emph{training} distribution)
%--- as expected, there is little overfitting going on.
The first four plots in Figure~\ref{fig:example} show the test 
performance using the area
under the ROC metric on the training distribution (the dashed red
curve) and on the test distribution (the solid red curve) vs. the
number of features we have greedily added.  Note two striking effects:
1) how rapidly the test performance plummets; 2) more troubling, how
quickly the test performance for the training and test distributions
diverge. In particular, note that the true generalization 
performance on the source 
training distribution is not at all
reflective of the true generalization 
performance on the target test distribution, 
even after adding just a few features:
%This last fact is essentially the classic issue of overfitting. 
a classic example of overfitting.
The final plot shows the average of the training and test performance,
averaged over which digit is held out, and cycling through
digits. 

%At some level, perhaps we should not be surprised as this experiment
Overfitting is to be expected, because this experiment 
violates the learning theoretic preconditions for successful
generalization. % --- though we still seek recompense. 
Furthermore, for
this particular experiment, we could argue that a generative approach
is more robust: if we have a model for generating a
``$2$'', then it should be good for recognition in diverse settings. While
the generative framework is promising, particularly for generating
predictive features, 
%often, empirical loss driven methods outperform
%them, and 
it is often difficult to specify good generative models.
% (like in
%our later experiments on learning similarity functions).

In this work, we assume a distribution over domains, and that our
training sample consists of a \emph{small} number of sample domains
independently drawn from the distribution over domains and where we have 
access to many samples in each domain. The goal in our setting is to perform well on
new domains sampled from this distribution. For example, in the
previous experiments, we can consider that we have eight sample  
(known) domains in our training set, where domains are of the form ``2
vs. 0'', ``2 vs. 1'', ``2 vs. 3'', etc.  This is much like
the standard supervised learning model, except that sampled ``points''
are now ``domains''.  The challenge is that we desire to avoid
overfitting with an extremely small number of domains, or few 
``samples'', compared to standard supervised
learning paradigms, that assume hundreds to millions of
samples.
 
% --- in
%particular, with fewer samples (e.g. fewer number of domains) than we
%are traditionally accustomed to using in our standard supervised
%learning paradigms, where we typically have hundreds to millions of
%samples.

The problem of domain adaptation is more general
than this particular formulation, where our focus is on how to do well on
a new \emph{random} domain. There are numerous different aspects of
the domain adaptation problem that have been studied. For example,
assumptions considered are: when the classes are ``imbalanced''
(e.g. when $\Pr[Y|D]$ could vary with the domain $D$); ``covariate
shift'' \cite{SchefferShift} where $\Pr[X|D]$ varies with the domain
$D$, while $\Pr[Y|X,D]$ is not a function of the domain; under a
change of representation, the joint distributions of $\Pr[(X,Y)|D]$ is
more similar~\cite{blitzer06,xue08,guo09,huangyates,adaptNLP}; settings where
one desires mixtures of predictors which adapt to each
domain~\cite{daume06}. A detailed discussion of these models is beyond
the scope of this paper (see \cite{JiangReview} for a more
comprehensive review.). There is also a growing body of theoretical
work on this problem, including~\cite{huang07,ben-david07,cortes08,baxter00}
that concentrates on either characterizing the degradation that can occur due to
distributional shift (e.g. ~\cite{ben-david07}) or robustly training
using biased sampling, such as the sample selection bias work of
~\cite{cortes08}.

Our work differs in that we assume a distribution over domains, and
our focus is on generalization on new domains.  One
interesting application of this work is on learning similarity
functions. For example, we may desire to learn a similarity function
for objects, where objects of the same label have high similarity, in
manner so as to be able to utilize this similarity function to
recognize new objects, not present in our training set; a problem known as 
``zero-shot'' learning. 
%Here, while train the similarity function
%using only a few object classes ($9$ classes in our experiments), we
%hope that it can generalize to objects not in our training set.

%\begin{figure}
%
%  NOTE:  These figures are built in the plots_from_R directory.  Look
%  in .figures for the most recent copy if you are building it
%  yourself. Or what Dean does is symbolic links that directory to here.
%
%
%\centerline{\includegraphics[width=5.75in,height=1.5in]{overfit-intro5ROC}}
%\vspace*{-.2in}
%\centerline{\includegraphics[width=7in,height=1.5in]{overfit-intro5tstats}}
%\label{fig:example}
%\caption{
%Area under ROC (AUROC) (y-axis) for predicting digit '2'
%using the greedy
% algorithm.
%The x-axis shows the order of variables
% were picked out of a list of 2000 total variables.   The top dashed
%line is the validation data.  The transfer to a new domain is
% shown much below it as the solid.  The last figure on the right shows
%the average of all the ROC curves. The horizontal line is at
%``chance.''} 
%\end{figure}

{\bf Our Contributions:\/} Our analysis focuses on the issue of
overfitting, and we borrow the idea from small sample statistics that
a certain empirical variance should be utilized when deciding
whether or not an effect is significant, namely, that an added feature will decrease
our error. We do this using $T$-statistics. The key idea is
that we can estimate the weight of each feature on each training
domain \emph{separately}. Indeed, if this weight varies wildly over the
training source domains, then even though this feature may be
useful on all our source domains, its potential for generalization
to new domains may be poor. We show that our data-dependent version of feature
selection robustly enjoys the usual feature selection properties,
i.e. we can select many more features than domains, particularly if
certain data-dependent variances are low, under relatively weak
assumptions.

The contributions of this work are as follows:
\vspace{-0.05in}
\begin{itemize}
  \item Using small sample statistics, namely that of $T$-tests, we
    provide a more robust procedure to add features, which takes
    into account data-dependent properties. 
\vspace{-0.02in}
  \item Using the theory of large deviations for self-normalized sums,
    we show that we can robustly add many more feature than domains
    (exponentially more), utilizing certain empirical variances. 
%To our
%    knowledge, these deviation bounds have not been utilized in
%    the analysis of  machine learning algorithms.
\vspace{-0.02in}
  \item We empirically demonstrate that we control for overfitting
    using an alternative greedy procedure for feature addition, based
    on the $T$-statistic. In particular, we show that these ideas can
    be utilized towards the theory of ``zero-shot learning''.
% (when
%    learning with similarities).
\vspace{-0.1in}
\end{itemize}

\section{Setting}
\vspace{-0.1in}
A key idea in our setting is that we consider a distribution over
domains, which we denote by $\Pr[D]$ (it is possible that there may be
an infinite number of domains). Conditioned on a domain $D=d$, the
distribution over input/output pairs is $\Pr[ (X,Y) | D=d]$, where
our inputs are $X\in \R^p$. %, and targets $Y \in R$.  
As is standard, these inputs could represent
a high dimensional feature space. The goal is to find a weight vector
which minimizes the squared error, averaged over both instances and
over the domains. More precisely, the error we want to minimize
is:
\[
\calL (w) = \Expct_{D} \Expct_{X,Y}[ (Y - w \cdot X)^2|D]
\]
where the inner and outer expectations are over $(X,Y)$ and $D$,
respectively. 

Our training set consists of a set of $n$ known domains $\{d_1,d_2, \ldots
d_n\}$, where each domain is sampled independently. In practice, while
$n$ is small, we often have a large number of samples in each domain,
so that the second order statistics can be estimated accurately on
each training domain. As a natural abstraction, we assume that for each domain
$d$ in our training set, we have knowledge of both $\E[XY|D=d]$ and
$\E[XX^\top|D=d]$.

For our theoretical analysis, we also assume the joint input
covariance matrix $\E[XX^\top]$ is known, as it can be estimated
accurately with unlabeled data. This permits a cleaner exposition
in terms of unbiased estimation, although this distinction is
relatively minor in practice.

\section{Feature Selection and Small Sample Statistics}
\vspace{-0.1in}
Our goal is to avoid overfitting while adding features: we desire
confidence that our added feature actually improves the error on
\emph{new} domains. The naive greedy method is to add features which
maximally decreases our training set error, which, as we have shown in
the Introduction, can perform remarkably poorly.  Instead, we provide
a theory which more sharply characterizes when adding a feature
actually improves our performance.

\subsection{Adding a Single Feature}
\vspace{-0.05in}

%Let us characterize the simplest question of whether or not a single
We first investigate the question of whether or not a single
feature improves the null prediction of always saying $0$. It is
natural to base our theory using unbiased estimates, as we often have
the most robust statistical tests for these estimates.

Consider a feature $X_i$, which is normalized so that
$\E[X_i^2]=1$. The optimal weight on this feature is $w^*_i = \E[X_i
Y]$.  Furthermore, any weight $w_i$ on $X_i$ has regret:
\[
\calL(w_i) -\calL(w^*_i) = (w_i - \E[X_i Y])^2
\]
Hence, with respect to adding just one feature, our task is to find a
feature $X_i$ and weight vector $w_i$ such that we have confidence that
$w_i$ is closer to $\E[X_i Y]$ than $0$ is (as weight $0$ corresponds
to the null prediction).
The natural unbiased estimate for $w^*_i$ is simply:
\[
\hmu_i = \frac{1}{n}\sum_{k=1}^n \E[X_i Y|d_k]
\]
The Central Limit Theorem implies that $\hmu_i$ should be close to
$\E[X_i Y]$ on the order of $O(\frac{\sigma(X_i Y)}{\sqrt{n}})$, where
$\sigma(X_iY)$ is the standard deviation.  A key idea in small sample
statistics is to take into account the empirical variance. Here, when
determining if $X_i$ is useful, we seek to consider the (unbiased)
variance estimate:
\[
\hsigma_i^2 = \frac{1}{n-1}\sum_{k=1}^n (\E[X_i Y|d_k] - \hmu_i)^2
\]
and the issue is how to utilize this estimate rather than the true variance.

In our domain adaptation setting, it may be the case that this covariance
for certain ``robust'' features $\E[X_i Y]$ is more consistently
correlated with the target --- it is these features that we seek to
add.  By contrast, ``large'' sample analysis typically involves only
using an upper bound on the standard deviation $\sigma(X_i Y)$, along
with tail bounds such as the Bernstein bound \cite{Bernstein46}, to
get estimates on the deviation between $\hmu_i$ and its
mean. However, crucially, as $\sigma(X_i Y)$ could vary greatly with
our feature $X_i$, we desire a sharper estimate which takes into
account the empirical variance, $\hsigma_i^2$.

If $\hmu_i$ followed a normal distribution, then this question
reduces to a Student's $T$-test. Here, the $T$ statistic is:
\[
T_i = \frac{\hat \mu_i}{\hat\sigma_i / \sqrt{n}}
\] 
While we do not expect the $\hmu_i$ to actually follow a normal
distribution, there is a rather large literature showing that the
$T$-test is robust (see for example ~\cite{SelfNormBook}).  We
now demonstrate this point under a milder assumption, that $\hat
\mu_i$ is symmetric (where the source of randomness is from a random domain).
Equivalently, this is an assumption that the
covariances $\E[X_i Y|d]$ are symmetric about their mean (i.e. both
$\E[X_i Y|d] - \E[X_i Y]$ and $-(\E[X_i Y|d] - \E[X_i Y])$ have the same
distribution, where $d$ is the source of randomness). The following
theorem assumes \emph{no} moment conditions on $X_i$ or $Y$ (not even
upper bounds). It shows
that we can accurately test an exponential number of features with
high confidence. This bound has similar behavior to the
$T$-distribution (for fixed $n$) as we scale the number of features.

\begin{theorem}\label{thm:one}
  Assume the random vector $\E[X Y|d] - \E[X Y]$ is symmetric (where $d$
  is random).  Let $\delta>0$. Suppose $\calF$ is a set
  of features whose size satisfies $|\calF| \leq \frac{\delta}{2}
  e^\frac{n}{8}$ (e.g. it is of size at most exponential in $n$).  Then for all
  $X_i$ in $\calF$, we have with probability greater than $1-\delta$:
\[
|\hat\mu_i - \E[X_i Y]| \leq \frac{\hsigma_i}{\sqrt{n}}
\  \sqrt{4 \log \frac{2 |\calF|}{\delta}}
\]
where \emph{no} moment bounds on $X$ and $Y$ are assumed,
aside from existence of $\E[X Y|d]$ and $\E[XY]$.
\end{theorem}

The proof of this theorem is in the appendix. The key is that this
theorem shows that the empirical variance can be taken into account
when searching through a large feature set.  
%Also note that this bound
%compares favorably well with the idealized case in which the random
%variables $X_i Y$ are IID normal (which can be explicitly
%verified). 
In fact, asymptotically, as implied by the Central Limit
Theorem, the only improvement possible is that the constant of $4$
would become a $2$.

The proof (provided in the Appendix) of this bound is significantly
more subtle than the standard ``Bernstein''-like bounds,
since the $T$-statistic has much ``thicker'' tails. Our proof is based
on the following bound for ``self-normalized'' sums, which, to our
knowledge, has not been utilized in the machine learning literature.

\begin{theorem}\label{thm:self}
  (See Theorem 2.15 in ~\cite{SelfNormBook}) Assume $Z_1,\ldots Z_n$ are independent,
  mean $0$, symmetric random variables.  For all $t>0$, the
  following bound on the self-normalized sum holds:
\[
\Pr\left[ \  \frac{ (\sumin Z_i)^2}{ \sumin Z_i^2} > t \ \right] \leq \exp\left(-\frac{t}{2}\right)
\]
where no moment bounds on $Z_i$ are assumed, aside from its mean existing.
\end{theorem}

For completeness, we add the proof of this theorem in appendix. It is
based on a simple symmetrization argument along with Hoeffding's tail
inequality. Note that the above bound is not quite a large deviation
bound for a $T$-statistic, as the denominator uses $\sumin Z_i^2$,
while a $T$-statistic would have a term of the form $\sumin (Z_i-\hat
Z)^2$, where $\hat Z$ is the empirical estimate of the mean, $\sumin
Z_i/n$.  This subtlety leads to the condition in
Theorem~\ref{thm:one} that the size of $\calF$ is at most 
exponential in $n$.

\subsection{Subset Selection}
\vspace{-0.05in}
Merely searching for the lowest error solution over all subsets of, say,
size $q$ is prone to overfitting. Instead, we seek to take into account
the empirical variance when searching over subsets of features.  We
now provide a data-dependent bound showing that the empirical variance
can be utilized for a much sharper bound. In the next subsection, we
discuss a greedy method for this search.

Given some set of features $\calS$ of size $q$, let $\basisX_1 \ldots
\basisX_q$ be an orthonormal basis for this subspace
(e.g. $\E[\basisX_i \basisX_j]$ is $0$ if $i\neq j$ and $1$ if
$i=j$. Note that we can put $\calS$ into this basis as we have assumed
knowledge of $\E[XX^\top]$). The best weight vector for this subspace
%(in this basis) 
is again the covariance $[\mu_S]_i=\E[\basisX_i Y]$. Define the
(unbiased) empirical means and variances as follows:
\beqa
[ \hmu_S]_i &=& \frac{1}{n}\sum_{k=1}^n\E[\basisX_i Y|d_k], \nonumber \\
\textrm{\tiny{ }} [\hsigma_S]_i^2 &=& \frac{1}{n-1}\sum_{k=1}^n (\E[\basisX_i Y|d_k] - [ \hmu_S]_i)^2
 \nonumber 
\eeqa
We take $\hmu_S$ as the estimate of the weight vector on this subspace.
We now provide our data dependent generalization bound, in terms of an appropriate
empirical variance. In particular, we are interested in a generalization bound
for all subsets of size $q$ out of a feature set of size $p$.

\begin{corollary}~\label{corr:subsets}
  Assume the random vector $\E[X Y|d] - \E[X Y]$ is symmetric.
  Let $\delta>0$. Assume that our set of features $\calF$ is of size
  $p$, and that $qp^q \leq \frac{\delta}{2} e^\frac{n}{8}$.  For
  all subsets $\calS\subset \calF$ of size $q$, we have:
\[
L(\hmu_S) - L(\mu_S) \leq 
\left( \sum_{i \in \calS} [\hsigma_S]_i^2 \right) 
\frac{8q \log p + \log (2/\delta)}{n}
\vspace{-0.1in}
\]
\end{corollary}

This bound is analogous to the usual bounds for regression where
instead of the sum empirical variance, we have the true variance
(which is usually assumed to be constant in idealized Gaussian noise
regression model\footnote{For the usual model, where $Y=\beta X
  +\eta$ where $\eta$ is Gaussian noise with variance $\sigma^2$. The
  risk bound above is just $\sigma^2 \frac{q \log p}{n}$, which
  is improved by a factor of $q$. We conjecture if we made the further
  assumption that the random vector $\E[X_i Y|d] - \E[X_i Y]$ is
 spherically symmetric, then the factor of $q$ can be removed.}). Crucially, the
bound shows that we can robustly utilize the empirical variance when doing our
estimation.
The implications of this are that we can design a much sharper
procedure for testing if a feature improves performance.

\subsection{Practice: The T-Greedy Algorithm}
\vspace{-0.05in}
In practice, the natural methodology is to ``greedily'' choose a
feature instead of searching all subsets, which usually consists of
finding the feature which decreases the error the most. Instead, we
introduce the $T$-greedy algorithm, a ``stagewise'' procedure for  adding the feature which has the
highest $T$-statistic (e.g. the goal is to add a feature in which we have the most
confidence that the true error will be improved).

There are a variety of greedy regression procedures, such as
``stepwise'',``stagewise'', etc. We now present a stagewise variant
by considering covariances with our residual error $(Y-w \cdot X)$.
Suppose that our current weight vector is $w$ (on our current set of features).
For each feature $X_i$, we compute the empirical mean and variance:
%of the covariance of the feature with the residual, e.g.
\beqa
\hmu_i &=& \frac{1}{n}\sum_{k=1}^n \E[X_i (Y-w\cdot X)|d_k], \nonumber \\
\hsigma_i^2 &=& \frac{1}{n-1}\sum_{k=1}^n (\E[X_i (Y-w\cdot X)|d_k] - \hmu_i)^2
\nonumber
\eeqa
Note that with a finite number samples in each domain, we would simply use the
empirical estimates instead. Now we just add the feature with the
highest $T$-statistic, e.g. add the feature:
\[
i^\star = \argmax_i T_i
\]
where $T_i = \frac{\hat \mu_i}{\hat\sigma_i / \sqrt{n}}$. Now our
update to the weight on this feature is simply:
\[
w_{i^\star} \leftarrow w_{i^\star}+ \frac{ \hmu_i }{ \hat \E[X_i^2]} ,
\quad \hat \E[X_i^2] = \frac{1}{n}\sum_{k=1}^n \E[X_i^2|d_k]
\]
Observe that this is actually a biased estimate of the optimal weight on
this added feature. Technically, our theory is only applicable to
using unbiased estimates, where we would have $\E[X_i^2]$ in the
denominator. This is a minor distinction in practice, and with
unlabeled data we could essentially run the unbiased version. We
should point out that stepwise variants are also possible.

\begin{figure*}[t!]
\vspace{-0.35in}
\hbox{ \centering \hspace{-0.05in}
\centerline{\includegraphics[width=5.9in,height=1.75in]{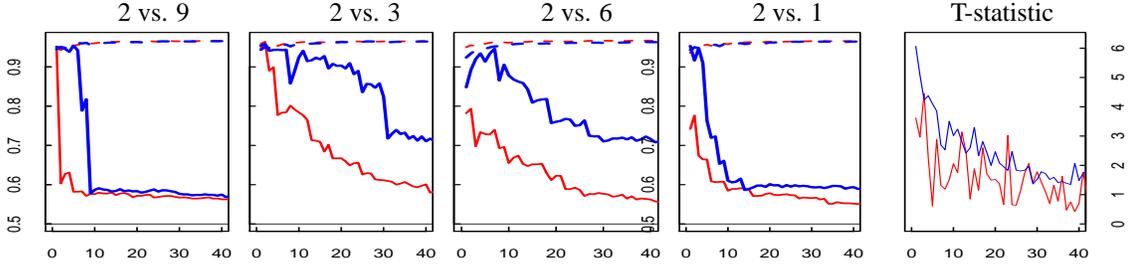}}
%\setlength{\epsfxsize}{6.0in}
%\epsfbox{figures/overfit-newTWO}
\begin{picture}(0,0)(0,0)
\setlength{\epsfxsize}{1.7in}
\put(-380, +108){ 2 vs. 9}
\put(-300, +108){ 2 vs. 3}
\put(-220, +108){ 2 vs. 6}
\put(-140, +108){ 2 vs. 1}
\put(-65, +108){ T-statistic}
\end{picture}
}
\vspace*{-.3in}
\caption{ \small
 \label{fig:mnist2} Area under ROC (AUROC) (y-axis) for predicting digit '2'
 for both the greedy
 algorithm (shown in red) and for our $T$-greedy algorithm (shown
 in blue), as we add more features (x-axis).  Since we
 choose features based on T-statistics (see last graph), our T-statistics is generally higher
 than that of the greedy algorithm.}
\vspace{-.1in}
\end{figure*}

\iffalse
For example, if our
current set of features is $\calS$, when deciding to add the feature
$X_i$, we first orthogonalize $X_i$ to those features in $S$ to obtain
a feature $\basisX_i$. Then we look at the $T$-statistic of the
variable $\basisX_i Y$, and we choose the feature with the highest $T$
statistic.

\begin{algorithm}[t]
{ \small
\begin{algorithmic}[1]
\STATE Given: $S$ is a set of features, $V$ is empty. 
\FOR {$t=1:T$ (number of iterations) }
\STATE Compute T-statistics $\mu/\sigma$. 
\STATE
 Pick feature $X_i$ from $S$ with highest T-statistic.
 \STATE Add $X_i$ to $V$. 
\STATE Fit parameter $w_i$ 
\ENDFOR
\end{algorithmic}
\caption{T-greedy algorithm. }
 \label{alg:T-greedy}
}
\end{algorithm}
\fi

\iffalse
In terms of determining when to stop, we could either do a statistical
test checking for confidence that we improving (appropriately making a
Bonferoni correction (e.g. a union bounded) due to checking many
features). Alternatively, we could use a cross validation procedure
(over the domains) to determine when to stop.
\fi

\section{Experimental Results}
\vspace{-0.1in}
We now present results on the MNIST and CIFAR image datasets.  The
MNIST digit dataset contains 60,000 training and 10,000 test images of
ten handwritten digits (0~to~9), with 28$\times$28 pixels.  In all
experiments, we use 10,000 digits (1,000 per class) for training and
10,000 digits for testing.  Instead of using raw pixel values, each
image was represented by 2000 real-valued features, that were
extracted using a deep belief network \cite{HOT}.

We also present results on the more challenging CIFAR image dataset
\cite{Cifar}, that contains images of 10 object categories, including
airplane, car, bird, cat, dog, deer, truck, deer, frog, and horse.
As with the MNIST dataset, we use 10,000 images (1,000 per class)
for training and 10,000 images for testing.  Each image was also
represented by 2000 real-valued features, that were extracted using a
deep belief network \cite{Cifar}. We note that extreme variability in
scale, viewpoint, illumination, and cluttered background, makes object
recognition task for this dataset difficult.

In all experiments, we report the area under ROC (AUROC)
metric of two different algorithms,  that we refer to as the {\it greedy} and
our proposed {\it T-greedy} algorithm.
The greedy algorithm %is the linear stepwise regression
chooses the next feature which decreases the
squared loss the most on the training set.
The T-greedy algorithm, on the other hand, % is the linear stepwise regression
chooses a feature with the largest T-statistic.
For both methods, we report both
generalization error on our training or 'source' domains
as well as generalization error on test or 'target' domains.  We do
not focus on the issue of stopping but rather on robustness. There
are a variety of methods for stopping which we mention in the
Discussion section.

\begin{figure*}[t!]
\vspace{-0.3in}
\hbox{ \centering \hspace{-0.05in}
\centerline{\includegraphics[width=5.9in,height=1.75in]{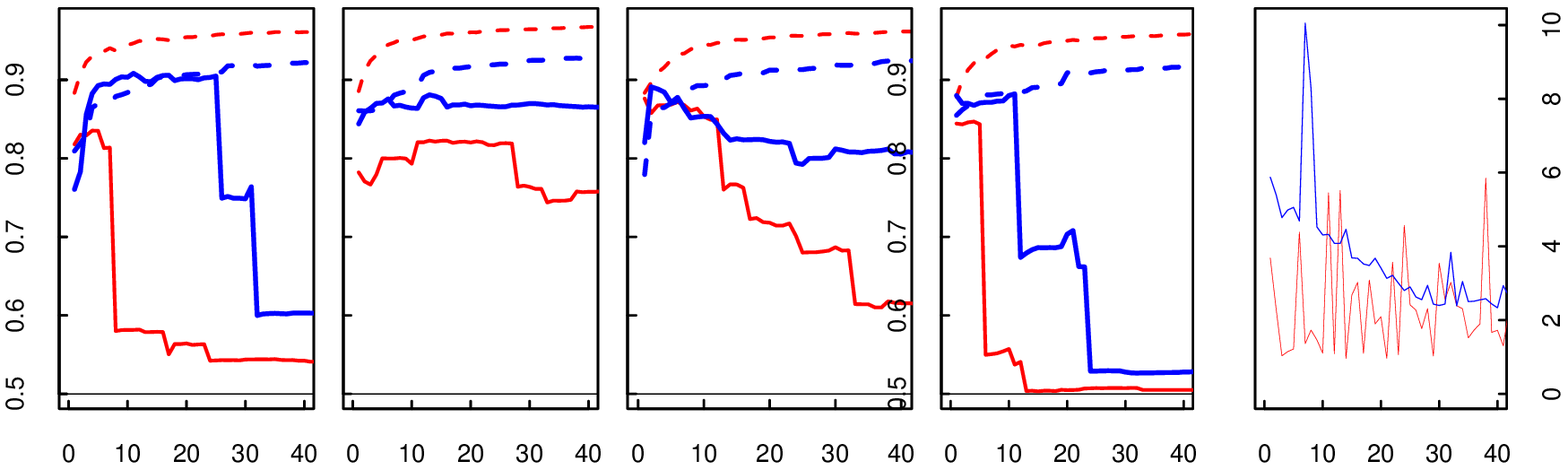}}
\begin{picture}(0,0)(0,0)
\setlength{\epsfxsize}{1.7in}
\put(-380, +108){ nine}
\put(-300, +108){ three}
\put(-220, +108){ six}
\put(-140, +108){ one}
\put(-65, +108){ T-statistic}
\end{picture}
}
\vspace{-0.3in}
\hbox{ \centering \hspace{-0.1in}
\centerline{\includegraphics[width=5.9in,height=1.75in]{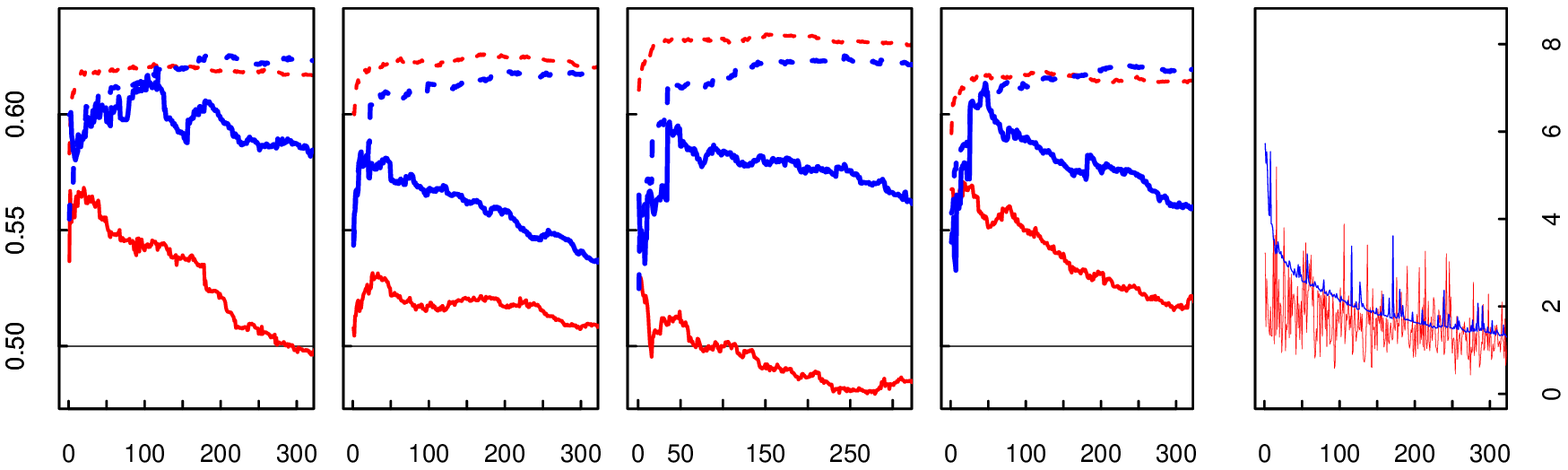}}
\begin{picture}(0,0)(0,0)
\setlength{\epsfxsize}{1.7in}
\put(-380, +108){ airplane }
\put(-300, +108){ bird}
\put(-220, +108){ cat}
\put(-140, +108){ truck}
\put(-65, +108){ T-statistic}
\end{picture}
}
\vspace*{-.35in}
\caption{\small
 \label{fig:mnist_sim}
Area under ROC (AUROC) (y-axis) for learning similarity function
 for both the greedy
 algorithm (shown in red) and for our $T$-greedy algorithm (shown
 in blue), as we add more features (x-axis).  Since we
 choose features based on T-statistics, our T-statistics are generally higher
 than that of the greedy algorithm.
 }
\vspace*{-.15in}
\end{figure*}

\subsection{MNIST (2 vs. other)}
\vspace{-0.05in}
In our first experiment, shown as the leftmost plot in
\Figref{fig:mnist2}, we tested the ability of the proposed algorithm
to generalize to a new target domain: recognizing the 
digit '2' vs. the new, previously unseen digit '9'.
To this end, we created eight source domains: \{'2' vs. '0'\},...,
\{'2' vs. '8'\}, where each domain contained a balanced set of 2000
labeled training examples\footnote{Remember, our key assumption 
is that the sampled domains are independent and that the source 
domains are known.}.
Our new target domain (as the test set) \{'2' vs. '9'\} also
contained a balanced set of 2000 examples.

\Figref{fig:mnist2}, the leftmost plot, displays an evolution of the area under ROC (AUROC)
metric for both greedy (red curves) and $T$-greedy (blue curves) algorithms.  
Note that an area of 0.5 corresponds to a random classifier, shown on
the graph as a horizontal line.
The dashed curves correspond to generalization or 'test' 
performance on the source domains, whereas the solid curves 
display performance on the target domain. 
Observe that after adding a few features, 
test performance of the greedy algorithm on the new 
target domain (red solid curve) rapidly decreases.
Test error on the source domains, however, keeps improving,
clearly demonstrating that no overfitting on the source domains is occurring. 
Hence, for the greedy algorithm, the true 
error on the source and target domains rapidly diverge.
% The following
%three plots show this for target domains of \{'2' vs. '3'\}, \{'2'
%vs. '6'\}, and \{'2' vs. '1'\}.
%Even worse, after adding only a single features, we already see a huge gap
%in the true        
%error on the source and target domains (difference in red-dashed
%and red-solid curves).  

This is in sharp contrast to the performance of the $T$-greedy
algorithm.  Even though performance of the $T$-greedy algorithm on the
source domains (blue dashed curve) is slightly worse (as expected as
it is not as aggressively striving for source error minimization), the
true AUROC on the source and target domains diverges less rapidly ---
in particular, these curves start close together.  \Figref{fig:mnist2}
further shows results for different source/target splits.  We
consistently observe that as we add few features, the $T$-greedy
algorithm overfits much less on the target domain.  This consistency
is also seen in left most plot of \Figref{fig:avg}, that displays 
results averaged over all splits of the source and target domains. 
%.  But the
%average plot loses the variability that in inherent in the problem:
%each domain behaves very differently than the others.

The rightmost plot of \Figref{fig:mnist2} also 
shows the $T$-statistic of the added feature to the
model of both algorithms.  We only show one such figure since they all
look similar.

We formulate the similarity learning problem in our regression setting as follows.
Given two feature vectors corresponding to two
images $\phi(X^1)$ and $\phi(X^2)$, we consider a linear regression function:
%\vspace{-0.05in}
\beqa
   y = \textrm{sgn} (\sum_i w_i \phi_i(X^1) \phi_i(X^2)), \nonumber
\eeqa
where we set $y=1$ if two images have the same label (positive example),
and $y=-1$ if two images have different labels (negative example).

\Figref{fig:mnist_sim}, top row, displays results on learning a similarity function
for MNIST digits. In particular, consider learning a similarity
function on all the
digits, but with digit '9' excluded. Similar to the previous experiment,
we constructed nine source domains (corresponding to digits 0 through 8).
Each domain
contained 1000 positive and 1000 negative examples, where negative examples were
randomly sampled from the remaining digits in the source domain.
Our target domain contained 1000 positive examples of newly observed images of '9'
and 1000 negative examples, randomly sampled from images of 0-to-8.

\subsection{Learning similarity function}
\vspace{-0.05in}
We now consider a more demanding task of
learning a similarity function between two images.
A good similarity function can provide insight into
how high-dimensional data is organized and
can significantly improve
the performance of many machine learning algorithms
that are based on computing similarity metric.
Our goal
%The key difference between our work and much of the existing work \cite{}
is to learn a similarity function that can not only work well for
objects that are part of the training set, but also works well for new
objects that we may have never seen before: a widely studied problem known as a
``zero-shot'' learning.

\begin{figure*}[t!]
\vspace{-0.1in}
\hbox{ \centering \hspace{-0.05in}
\centerline{\includegraphics[width=5.9in,height=1.75in]{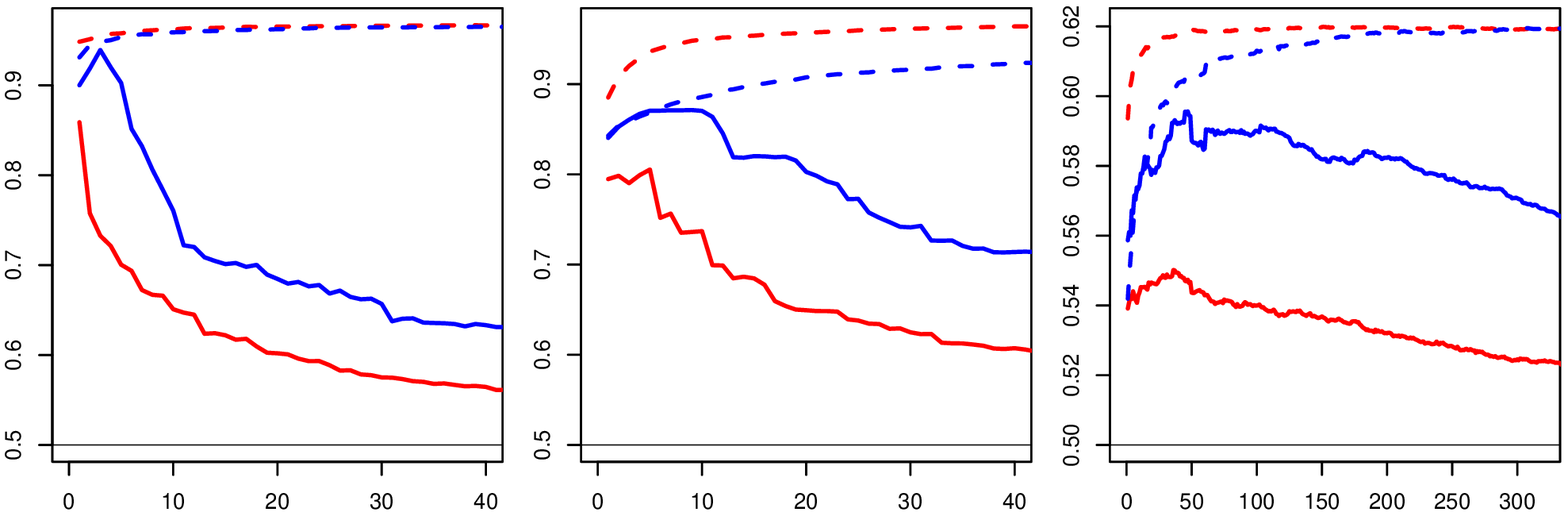}}
\begin{picture}(0,0)(0,0)
\setlength{\epsfxsize}{1.6in}
\put(-380, +112){ Recognizing digit '2'}
\put(-235, +112){ MNIST similarity}
\put(-100, +112){ CIFAR similarity}
\end{picture}
}
\vspace*{-.3in}
\caption{ \small
 \label{fig:avg} 
  Area under ROC averaged over all splits of source/target domains for
 predicting digit 2 (left), learning similarity function for MNIST digits (middle)
 and CIFAR images (right).
 }
\vspace*{-.2in}
\end{figure*}

\Figref{fig:mnist_sim}, top leftmost plot, shows that 
the generalization error of the greedy algorithm  
on the source and target domains rapidly diverge. The $T$-greedy, on the 
other hand, is able to select up to 25 reliable features that help
us generalize well to the new target domain. 
\Figref{fig:mnist_sim} further displays performance results when 
generalizing a similarity function to different target domains.
Again, the rightmost plot shows the value of the added
$T$-statistic for one of these plots.

Finally, we experimented with learning a similarity function for more
challenging image CIFAR dataset.  Similar to the results on the MNIST
dataset, \Figref{fig:mnist_sim}, bottom row, shows that the $T$-greedy
algorithm is able to consistently pick up to 50 robust features that are
useful for transfer to a new domain (note the difference in scale on
the $x$-axis, which now goes to 300 features). The greedy algorithm,
however, barely improves upon making random predictions.

We have focused attention on the individual domains to help drive home
 how variable each domain is from the others.  But, it is sometimes
 hard to see the signal amongst all this noise, so we also provide
 averaged versions of the AUROC curves (\Figref{fig:avg}).  The
 $T$-greedy algorithm is able to pick up many more robust features and
 overfits far less on the target domain (difference in blue-dashed and
 blue-solid curves).  The greedy algorithm's test error diverges after
 adding only a handful of features.  Almost immediately we see a big gap in
 the error on the source and target domains (difference in red-dashed
 and red-solid curves).

\section{Discussion}
\vspace{-0.05in} All experiments demonstrate that the $T$-greedy
algorithm has better correspondence between training AUROC and testing
AUROC.  The curves start out with the training and the testing AUROC
curves with about the same value.  This is particularly striking in
the averaged curves, shown in \Figref{fig:avg}. So by looking only
at the training curves one can get a good estimate of the
generalization performance. As expected, eventually overfitting
occurs, since the training AUROC continues to improve whereas the
testing AUROC decreases.  However, even then it is possible to get a handle on
using our method (e.g. when to stop).  One option
is to simply keep yet another domain held out for cross-validation
and cycle through. 
% Since our curves are fairly smooth, hitting the
%maximum point is not critical (so we do not a very exact estimate of
%the best stopping point). 
Alternatively, we can use properties of the
T-statistic to get a handle on when to stop (e.g. when the
$T$-statistics is behaving like chance).  Here,  Bonferoni can also be
used as a heuristic to decide how many variables to use.  Again, this
is made easier by the fact that the curves are close.
% Neither of these will be pursued further in this
%paper. 

%\dpf{new summary paragraph about variability of figures}

We also observe that %The other key take home message is that
the variability between domains is much greater than the variability
within any given domain (Figs.  2, 3, and 4 all show this
variability).  Classical statistics assumes that each error is
independent (if just merged across all the domains), but we see from plots
that each domain behaves idiosyncratically.  Sometimes they overfit
after a few variables, sometimes they continue to improve.  This means
that using more observations from the domains we have already studied
is not informative of how we will extrapolate to new domains.  Such
small sample sizes were the original motivation for Gosset to come up
with his Student's T-statistic. Note that we do not have many degrees
of freedom but we can still obtain as much information out of the data
we have.  What we see from our analysis and experiments is that this
information can still be substantial.
{\small
\paragraph{Acknowledgments}
RS is supported by NSERC, and NTT Communication
Sciences Laboratory.
}

%\begin{figure}
%\centerline{\includegraphics[width=6in,height=1.75in]{figures/overfit-digit}}
%\vspace*{-.4in}
%\centerline{\includegraphics[width=6in,height=1.75in]{figures/overfit-cifar}}
%\vspace*{-.4in}
%\centerline{\includegraphics[width=6in,height=1.75in]{figures/overfit-grand}}
%\vspace*{-.2in}
%\caption{The first row show typical t-statistics followed by in and
%out of sample for the digits similarity problem. (Greedy shown in
%red and for our system shown blue).
% The second row is the same for the object classification problem.
%The third row show average AUROC curves for (1) the digits prediction
%problem, (2) the digits similarity problem and (3) the object
%classification problem.}
%\end{figure}

%\newpage
\bibliographystyle{plain}
\bibliography{transfer}

\newpage
\appendix

\section{Appendix}

First, let us prove Theorem~\ref{thm:self} (also, see Theorem 2.15 in ~\cite{SelfNormBook}).

\begin{proof} (of Theorem~\ref{thm:self}).  Let $\eps_i$ be Rademacher random variables (e.g. independent random variables which take values uniformly in $\{-1,1\}$). Since each $Z_i$ is symmetric, we have that the distribution of $ Z_i$ is identical to the distribution of $\eps_i Z_i$. Hence, we have that:
\[
\Pr\left[ \  \frac{ (\sumin Z_i)^2}{ \sumin Z_i^2} > t \ \right] =
\Pr\left[ \  \frac{ (\sumin \eps_i Z_i)^2}{ \sumin Z_i^2} > t \ \right]
\]
Now we bound this latter quantity for \emph{every} realization of $Z_i$.  
Consider a fixed set of values $z_1, \ldots z_n$ (some realization 
of $Z_1,\ldots Z_n$). 
For these fixed values, let us now bound the probability: 
\begin{eqnarray*} \Pr\left[ \ \frac{ (\sumin \eps_i z_i)^2}{ \sumin z_i^2} > t 
\ \right] \kern-6pt &=& \kern-6pt
  \Pr\left[ \  \left(\sumin \eps_i z_i\right)^2 > t  \sumin z_i^2 \ \right]\\
  \kern-6pt &=& \kern-6pt
  \Pr\left[ \  \left|\sumin \eps_i z_i\right| >  \sqrt{t \sumin z_i^2} \ \right]\\
  \kern-6pt &\leq& \kern-6pt
  2 \exp\left( -\frac{t \sumin z_i^2}{2\sumin z_i^2} \right) \\
    \kern-6pt &=& \kern-6pt
 2 \exp( -t/2) 
\end{eqnarray*} 
where the second to last step is by Hoeffding's inequality (where the only randomness is due to the $\eps_i$). To see this, note that we are adding the independent variables $\eps_i z_i$ which are mean $0$ and bounded in magnitude by $z_i$.  
\end{proof}

Now we prove Theorem~\ref{thm:one}.

\begin{proof} (of Theorem~\ref{thm:one})
For symmetric, mean $0$, independent $Z_i$, define:
\[
\hmu = \frac{1}{n}\sumin Z_i, \quad \hsigma^2 = \frac{1}{n-1}\sumin (Z_i - \hmu)^2
\]
The $T$-statistic is then defined as:
\[
T = \frac{\hmu}{\hsigma/\sqrt{n}}
\]
Define the related quantity:
\[
\widetilde T = \frac{\hmu}{\left(\frac{1}{n}\sumin Z_i^2\right)^{1/2}/\sqrt{n}} 
\]
and note that:
\[
\frac{ (\sumin Z_i)^2}{ \sumin Z_i^2} = \widetilde T^2
\]
Also, one can show that $\widetilde T^2 = \frac{n}{n-1}
\frac{T^2}{1+\frac{T^2}{n-1}}$. Hence, using the bound on
self-normalized sums,
\begin{eqnarray*}
\Pr[T^2 \geq t^2 ] & = & \Pr\left[\widetilde T^2 \geq \frac{n}{n-1} \frac{t^2}{1+\frac{t^2}{n-1}}\right]\\
 & \leq & 2 \exp\left(-\frac{1}{2} \frac{n}{n-1}
   \frac{t^2}{1+\frac{t^2}{n-1}}\right)\\
 & \leq & 2 \exp\left(-\frac{1}{2} \frac{t^2}{1+\frac{t^2}{n-1}}\right)\\
\end{eqnarray*}
Now let us choose $t=\sqrt{4 \log \frac{2 |\calF|}{\delta}}$. By
assumption on the size of $\calF$, we have that $t^2 \leq n/2$, and
so $\frac{t^2}{n-1} \leq \frac{n}{2(n-1)} \leq 1$ (since $n\geq 2$). Hence, 
\[
Pr\left[ T^2 \geq 4 \log \frac{2 |\calF|}{\delta} \right] \leq
2 \exp\left(-\frac{1}{2} \frac{t^2}{2}\right) =
\frac{\delta}{|\calF|}
\]
Our result now follows by the union bound (over all $|\calF|$ features).
\end{proof}

Our corollary now follows:

\begin{proof} (of Corollary~\ref{corr:subsets})
For any subset, the regret is:
\[
\sum_{i\in \calS} (\hmu_i - \E[\widetilde X_iY])^2
\]
Now note that there are no more than $p^q$ possible subsets. Also,
each subset comes with its own basis. So let us demand confidence on
all $qp^q$ possible basis elements. So we use Theorem~\ref{thm:one}
with a set of size $qp^q$ features (note that the $\log$ of the size
of this set is bounded by $2q \log p$). Our theorem now follows by
summing over the errors.
\end{proof}

%\pagebreak

\end{document}